\documentclass{article}

\usepackage[final]{corl_2026} 

\usepackage{amsmath,amsfonts}
\usepackage{algorithm}
\usepackage{graphicx}
\usepackage{algpseudocode}
\usepackage{amssymb}
\usepackage{bm}
\usepackage{booktabs}
\usepackage[table]{xcolor}
\definecolor{ethblue}{RGB}{33, 92, 175}
\hypersetup{
  colorlinks,
  citecolor=ethblue,
  linkcolor=ethblue,
  urlcolor=ethblue}
\newcommand{\grayrow}{\rowcolor{gray!10}}
\usepackage{siunitx}
\usepackage{caption}
\usepackage{wrapfig}

\usepackage{xcolor}
\usepackage[absolute]{textpos}

\newcolumntype{C}[1]{>{\centering}m{#1}}

\title{Approximate Imitation Learning for Event-based \\ Quadrotor Flight in Cluttered Environments}

\author{
   Nico Messikommer, Jiaxu Xing, Leonard Bauersfeld,  \\ 
   \textbf{Marco Cannici, Elie Aljalbout, Davide Scaramuzza} \\
   Robotics and Perception Group, \\
   University of Zurich, Switzerland. \\
   \texttt{[nmessi, jixing, bauersfeld, cannici, aljalbout, sdavide]@ifi.uzh.ch}
}

\begin{document}

\definecolor{somegray}{rgb}{0.5, 0.5, 0.5}
\newcommand{\darkgrayed}[1]{\textcolor{somegray}{#1}}
\begin{textblock}{8}(4, 0.7)
\begin{center}
\darkgrayed{This paper has been accepted for publication at the \\
Conference on Robot Learning (CoRL), Austin 2026}
\end{center}
\end{textblock}

\makeatletter
\let\@oldmaketitle\@maketitle
\renewcommand{\@maketitle}{
    \@oldmaketitle
    \begin{center}
        \centering
        \vspace{-15pt}
        \includegraphics[width=\textwidth]{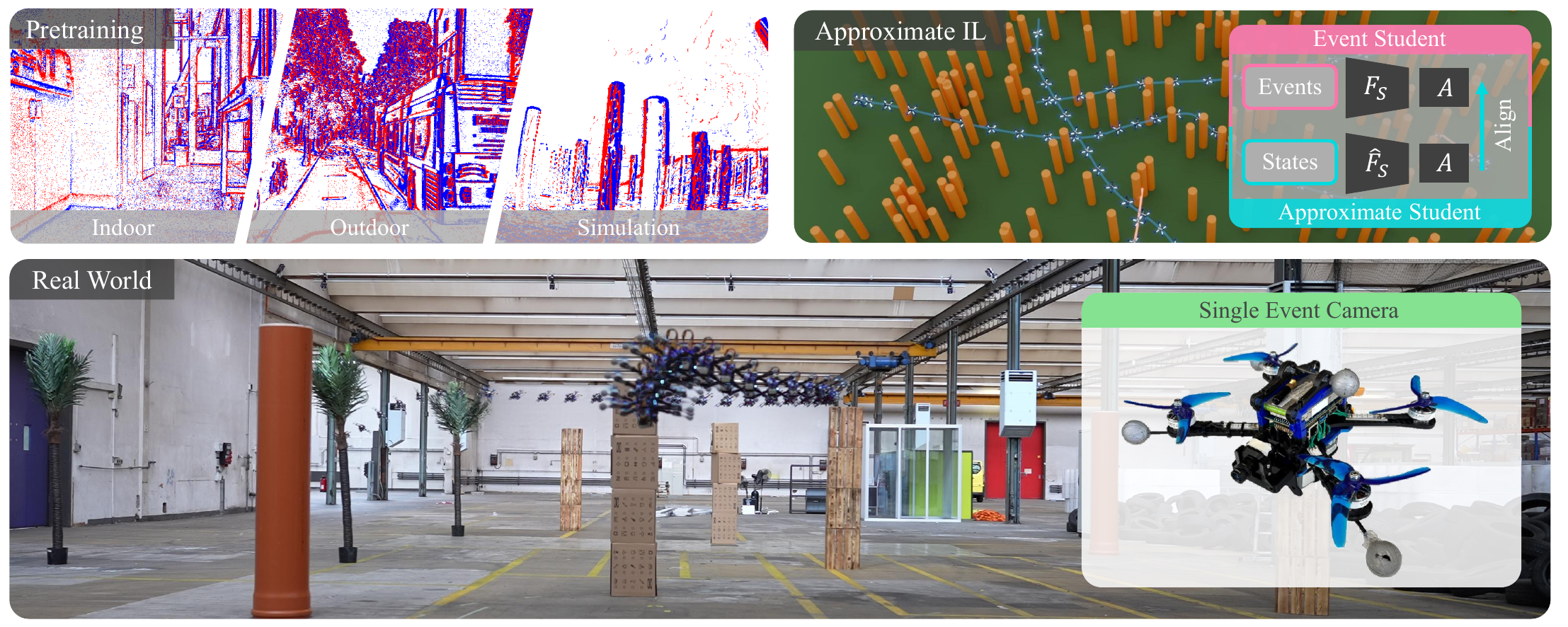}
        \vspace{-15pt}
        \captionsetup{type=figure}
        \caption{
        To fine-tune the event-based policy pretrained on offline datasets, we introduce an approximate student that receives simulated state information, eliminating the need for computationally expensive event rendering.
        We validate our framework by flying a quadrotor equipped with a single event camera through cluttered real-world environments using external pose information.
        }
        \label{fig:eyecatcher}
        \vspace{-7pt}
    \end{center}
    \bigskip 
}
\makeatother

\maketitle

\begin{abstract}
Event cameras offer high temporal resolution and low latency, making them ideal sensors for high-speed robotic applications where conventional cameras suffer from motion blur. 
However, their widespread adoption in robot learning is severely bottlenecked by the computational cost of simulating high-frequency event data during online training. 
In this work, we present \textit{Approximate Imitation Learning}, a novel framework that fundamentally resolves this bottleneck, reducing policy training time for complex, agile drone flight from 52.44 hours to just 1.86 hours—a $28\times$ computational speedup. 
Our key insight is to separate representation learning from policy search. 
We first leverage a large-scale offline dataset to learn a task-specific representation space. 
Subsequently, the policy is fine-tuned through online interactions that rely solely on lightweight state information, completely eliminating the need to render events during the active policy search phase.
This training paradigm drastically reduces development overhead and enables event-based control policies to scale to complex environments. 
Furthermore, our approach eliminates the reliance on standard cameras or intermediate representations during deployment, mapping events directly to control commands. 
In simulation, our method matches or exceeds the performance of standard imitation learning baselines that require full online event rendering. 
Finally, we successfully validate the framework in the real world, demonstrating that a policy trained via this ultra-efficient paradigm enables a quadrotor to fly through highly cluttered environments at remarkable speeds of up to $9.8\text{ ms}^{-1}$.

\end{abstract}
\vspace{-8pt}

\noindent\textbf{Multimedia Material}
A video is available at \url{https://youtu.be/yI8wGTaoYHA}

\section{Introduction}
Standard cameras capture dense visual information of the environment, enabling impressive robot performance across a wide range of applications~\cite{cadena2017past, levine2016end, kendall2019learning}.
However, their performance degrades during high-speed motion due to motion blur~\cite {sun2021frames, wu2022perception}.
Since standard cameras integrate incoming light over a finite exposure time, any image displacement exceeding one pixel during this period results in blurred images.
Consequently, perception quality deteriorates significantly, which negatively affects the downstream robot policy.
Event cameras offer a radically different sensing mechanism inspired by the biological vision system~\cite{Gallego20pami}.
Instead of integrating the incoming light, each pixel in an event camera asynchronously triggers events if the change in intensity exceeds a given threshold.
The asynchronous triggering of events reduces sensing latency to submilliseconds, compared to standard cameras that capture visual information at fixed frame rates.
As a result, they provide sharp visual information even in fast motions with low latency.
Furthermore, event cameras excel in scenes with very large brightness variations since they feature a high dynamic range that exceeds \SI{120}{\dB}.
These properties make event cameras an ideal choice to capture scene information reliably in high-speed robotic tasks.
An important application of reliable perception during high-speed motion is autonomous quadrotor flight for search-and-rescue missions, where rapid navigation in cluttered environments can be life-saving.
Prior work using standard cameras has demonstrated promising results in such environments~\cite{Zhang_2025}, but remains limited by motion blur.
To improve reliability, recent work has explored mounting LiDAR sensors on quadrotors~\cite{ren2025safety}.
While already smaller in scale, LiDAR sensors still have a high cost and are heavier.
Finally, methods using event cameras to increase sensing robustness have been proposed~\cite{falanga2020dynamic, bhattacharya2024monocular}.
However, most of them use a standard camera and can additionally rely on an intermediate representation, which introduces computational overhead~\cite{bhattacharya2024monocular}.
In our work, we propose an end-to-end approach, which maps events directly to commands without any intermediate representation or standard cameras, see Fig.~\ref{fig:eyecatcher}.
This enables the method to directly leverage the high temporal resolution of the event camera for fast quadrotor flight.
During training, to avoid the high computational cost of rendering high-frequency event data, our method first learns a task-relevant representation space from large-scale offline event datasets. 
This learned representation is then used within a lightweight simulator to train the control policy via the action decoder.
Crucially, the learning of the behavior does not rely on rendering events, making it highly efficient.
The results in simulation confirm the effectiveness of our proposed \textit{approximate Imitation Learning (IL)} framework.
Our method achieves a success rate increase of 0.2 over the behavior cloning baseline and an increase of 0.7 to the DAgger~\cite{ross2011dagger} baseline.
By leveraging the approximate student, our approach significantly reduces training costs, accelerating policy training by a factor of 28 (from 52.44 hours to just 1.86 hours).
Finally, we validate our approach in the real world, where the quadrotor achieves speeds up to \SI{9.8}{\meter\per\second} using external state information.
Although we apply our method only to event data, the framework is broadly applicable and can easily be extended to other modalities, which are computationally expensive to simulate, e.g., tactile sensors, radar, or LiDAR.

\section{Related Works}

\textbf{Robot Learning}
Achieving agile control with limited real-world data remains a central challenge in robot learning. 
While reinforcement learning (RL) has demonstrated strong performance in simulation~\cite{haarnoja2024learning} and real-world control~\cite{hwangbo2019learning, andrychowicz2020learning, aljalbout2024role, xing2024multi}, it is often sample-inefficient and computationally expensive. 
Imitation learning (IL) mitigates these issues using expert demonstrations~\cite{chi2023diffusionpolicy, shah2023vint, fu2024mobile}, yet methods such as DAgger~\cite{ross2011dagger} still require substantial interaction. 
Recent work improves efficiency further through pretrained visual representations~\cite{r3m, vip, xing2024contrastive}. 
Building on these advances, we propose an approximate IL framework for training-efficient, low-latency quadrotor navigation.
\textbf{Event-Based Quadrotor Flight}
Event cameras are increasingly used in robotics because of their high temporal resolution and low-latency output~\cite{forrai2023icra, wang2022ral, delbruck2013fnins}, making them well-suited for agile flight~\cite{stroobants2022neuromorphic, stroobants2025neuromorphic, paredes2024fully, sanket2020icra, vemprala2021representation}.
Early efforts focused on handcrafted, low-latency pipelines for tasks such as obstacle avoidance, achieving control loop latencies as low as \SI{3.5}{\milli\second}~\cite{falanga2020dynamic}.
Spiking neural networks (SNNs) have been explored for their compatibility with event data, including line tracking on constrained platforms~\cite{vitale2021event}, real-time drone flight via ego-motion estimation~\cite{paredes2024fully}, and dynamic obstacle avoidance using event–depth fusion~\cite{zhang2023dynamic}.
In parallel, conventional deep neural networks have been applied to tasks such as high-speed gate detection in drone racing~\cite{andersen2022event}.
In~\cite{bhattacharya2024monocular}, an event-based policy is trained in simulation via teacher–student distillation.
While effective, their approach relies on approximate event renderings with limited temporal resolution during training.
In contrast, we first learn a task-specific representation space from offline data, and then rely on an approximate student that uses state information to simulate the event-driven behavior, avoiding event rendering entirely.
\textbf{Vision-Based Quadrotor Flight}
Autonomous quadrotor navigation in cluttered environments commonly relies on onboard visual sensing, without external localization~\cite{mohta2018fast, tranzatto2022cerberus}.
Traditional model-based pipelines combine visual-inertial odometry (VIO)~\cite{bloesch2015robust, forster2016svo} with separate planning and control modules~\cite{karaman2010incremental, bircher2016receding}.
While effective in structured settings, they still suffer from brittle state estimation, delayed reactions, and limited adaptability in dynamic or perceptually degraded settings.
An early RL approach uses VIO as input to a policy that predicts discrete velocity commands~\cite{sadeghi2016cad2rl}.
Subsequent methods~\cite{kaufmann2018DDR, loquercio2018dronet, shah2022gnm, shah2023vint, sridhar2023nomad} learn to predict high-level control commands directly from visual observations.
To achieve tighter perception–action coupling, more recent approaches instead predict low-level control commands directly from images.
Notably, the integration of VIO with a simulation-trained, motion-capture-finetuned RL policy surpassed world champion pilots in drone racing~\cite{kaufmann2023champion}, and subsequent works achieve comparable agility through purely vision-based RL, without any state estimator~\cite{geles2024demonstrating, xing2024bootstrapping, romero2025dream}.
In parallel, IL has improved sample efficiency and has enabled agile end-to-end flight~\cite{Loquercio2021Science, song2020flightmare, xing2024contrastive} by minimizing the discrepancy between a learned policy and expert demonstrations~\cite{ross2011dagger, torabi2018behavioral}. 
Nevertheless, these methods still require millions of simulated interactions, which is especially problematic for event-based vision, where rendering event streams is computationally costly. 
In this work, we address this limitation through our approximate imitation learning, reducing sample complexity and making event-driven agile flight more tractable.

\section{Methodology}
\begin{figure*}
\centering
\includegraphics[width=0.99\textwidth]{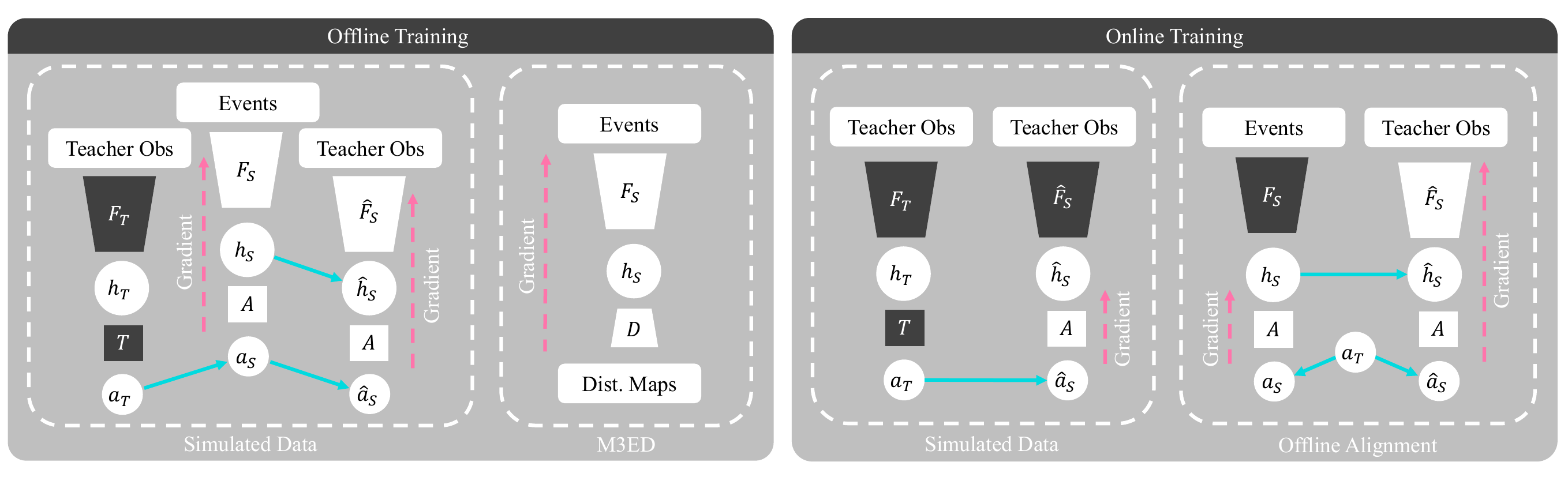}
\caption{
During offline training, teacher actions $a_T$ supervise the event encoder $F_S$ and the shared action decoder $A$.
The approximate student, using teacher observations, is trained by aligning its features $\hat{h}_S$ and actions $\hat{a}_S$ with the features $h_S$ and actions $a_S$ of the event student.
During online training, teacher observations obtained from lightweight simulations are used to fine-tune the shared action decoder $A$.
Consistency losses in both the action and feature spaces are applied during an additional offline alignment step.
For clarity, the auxiliary decoder $D_o$ is omitted.
}
\label{fig:approximate_IL}
\end{figure*}

In high-speed quadrotor flight through cluttered environments, the objective is to safely navigate obstacle-dense spaces while pushing the limits of perception by maximizing flight speed.
To instantiate this setting, the policy is conditioned on a commanded direction with a magnitude corresponding to the target velocity, which it tries to track while simultaneously avoiding obstacles.
To efficiently train an end-to-end policy using events, we adopt a two-stage learning approach that separates representation learning from policy search, as illustrated in Fig.~\ref{fig:approximate_IL}.
In the first stage, we train a mapping from events to a representation space using a pre-rendered offline dataset consisting of paired event representations and teacher actions.
To ensure that the learned representation space covers task-relevant information, we apply a Behavior Cloning (BC) objective based on the actions of the teacher.
We additionally incorporate a real-world dataset to train on an auxiliary depth task to improve generalization and robustness.
Nevertheless, BC struggles when the training data does not adequately cover the state distribution encountered during real-world execution.
To address this, we introduce a second stage that includes online student interaction in a simulator, crucially, without rendering events.
Instead, we use an approximate student model that observes full state information and learns to mimic the actions of the event-based student, which is inspired by~\cite{messikommer2025student}.
These interactions allow us to adapt the shared action decoder to a broader state distribution while avoiding the high computational cost of event rendering.
In the following, we provide a detailed explanation of each training stage.
The neural network architectures are introduced in the appendix.

\subsection{Offline Training}
\label{sec:offline_training}
The offline dataset is generated by rolling out the teacher policy for 40,000 steps across 25 parallel environments with small stochastic perturbations applied to the teacher actions.
Crucially, the expensive event rendering is performed only once and reused across multiple training runs.
The event-based student encoder $F_S$ is jointly trained with a shared action decoder $A$ using a standard BC objective to predict the teacher action $a_T$.
In parallel, we train an approximate student to match the latent features $h_S$ and action outputs $a_S$ of the event-based student using state observations.
Specifically, the approximate student is supervised with an L1 loss between the latent features of its encoder $\hat{F}_S$ and the event-based encoder $F_S$, and an MSE loss on the action predictions, both computed using the shared decoder $A$.
To reduce the sim-to-real gap~\cite{aljalbout2025reality} in the event-based student encoder $F_S$, we leverage the M3ED dataset~\cite{Chaney_2023_CVPR}, which includes event streams and depth maps collected from three platforms: a walking quadruped, a car, and a quadrotor.
Since teacher actions cannot be queried, we introduce an auxiliary depth prediction task.
Specifically, the student encoder is trained to predict 1-D angular distance maps, similar to a forward-facing single-line LiDAR scan restricted to the camera’s field of view, which resembles the teacher observations.
For the teacher, these distance maps are horizontally aligned using the gravity vector to compensate for tilt.
For the student, depth images are instead divided into three horizontal bands, and the minimum depth within \SI{11.25}{\degree} angular intervals is extracted for each band.
The three resulting distance maps are then concatenated into a single 1-D observation.
Using multiple bands reduces the need for gravity alignment by increasing the likelihood that at least one band captures an obstacle-free path, even when the camera faces the ground or sky.
These angular distance maps are predicted with a dedicated decoder $D$ using the features of the event encoder.
Similarly, using the rendered events in the offline dataset, the gravity-aligned distance maps that serve as input to the teacher are used to jointly train a separate auxiliary decoder $D_o$ and the event encoder $F_S$, further infusing task-relevant information.
The offline learning stage alternates between two update steps: a BC update and an auxiliary depth task update using simulated and real-world events from M3ED.
Each training iteration begins with a batch of real-world data, which is used to update the event-based student encoder $F_S$ and the distance decoder $D$.
This is followed by a batch from the simulated dataset, used to update the event-based student encoder $F_S$, the shared action decoder $A$, the approximate student $\hat{F}_S$, and the auxiliary decoder $D_o$.
Through this alternating update scheme, the learned representation is grounded in real-world event data and optimized for task relevance via supervision from expert demonstrations.
\subsection{Online Training}
To improve the generalization of the event policy, we further fine-tune the shared action decoder $A$ through online interactions in simulation.
Crucially, to avoid the computational cost of rendering events, we leverage the approximate student to approximate the behavior of the event-based student.
We adopt the DAgger framework~\cite{ross2011icais}, where both the teacher and the approximate student receive the same observations. 
An adaptive action sampling strategy, based on the difference between their predicted actions, gradually increases the proportion of actions taken by the student.
This setup enables the collection of diverse experiences for training the action decoder $A$ across a broader state distribution, while entirely bypassing the need for rendering events during online learning.
To prevent the action decoder $A$ from learning distinct manifolds for approximate–event student alignment and state-based behavior fine-tuning, we add an offline alignment update.
In this update, the action decoder is supervised to predict teacher actions from offline features generated by the event and approximate student.
In addition, the approximate student encoder is updated via feature consistency between the event and the approximate student, improving the robustness of the task decoder to small feature perturbations.
This alignment update step adds minimal overhead, since the event features $h_S$ for the offline data can be pre-computed once and reused for all fine-tuning runs.
\subsection{Event Rendering}
To improve render efficiency, we follow~\cite{bhattacharya2024monocular} and directly generate event representations, eliminating the computationally intensive step of synthesizing sparse event sequences.
In contrast to~\cite{bhattacharya2024monocular}, which generates single-channel representations, our proposed approach truthfully constructs a multi-channel event tensor that accumulates the events in a given temporal window for each polarity~\cite{mostafavi2021learning}.
Importantly, our method generates event representations for multiple videos in parallel using vectorized operations, significantly accelerating the simulation.
Given a high-frame-rate video with images $I_0, I_1, ... I_T$, we first compute the log-intensity for each frame and subtract the initial frame $I_0$, establishing it as the reference.
Each normalized log-intensity frame is floor-divided by the contrast threshold $C$, assigning each pixel at each timestep to a discrete \textit{contrast band}.
To detect band transitions, we compute the temporal difference of band indices between consecutive frames. 
Crucially, an event is only triggered when the intensity curve exits the current band in the opposite direction from which it previously entered.
This condition is met at timesteps where the second-order difference is zero, indicating that the logarithmic difference exceeded the contrast threshold $C$.
To obtain the final event representation, a simple reshaping and summing is required.
The event generation steps are summarized in Alg.~\ref{alg:event_representation}.
\begin{wrapfigure}{r}{0.48\textwidth}
\begin{minipage}{\linewidth}

\begin{algorithm}[H]
\footnotesize  
\caption{Event Representation Generation}
\label{alg:event_representation}
\hspace*{\algorithmicindent} \textbf{Input} Log frames $\{L_0, L_1, \ldots, L_T\}$, contrast $C$ \\
\hspace*{\algorithmicindent} \textbf{Output}  Event representation $E_T$
 
\begin{algorithmic}[1]
\State $\Delta L \gets \{L_t - L_{\text{0}} \mid t = 0, \ldots, T\}$  
\State $\text{BandID} \gets \lfloor \Delta L / C \rfloor$ 

\State $\Delta \text{BandID} \gets \text{BandID}_{1:T} - \text{BandID}_{0:T-1}$ 
\State $\text{BandID} \gets \text{NonZero(}\Delta \text{BandID)}$

\State $\Delta^2 \text{BandID} \gets \text{BandID}_{1:F} - \text{BandID}_{0:F-1}$
\State $E_T \gets \sum_{\text{axis}} \left[ \Delta^2 \text{BandID} = 0 \right]$

\end{algorithmic}
\end{algorithm}

\end{minipage}
\end{wrapfigure}
Except for a few corner cases related to specific initial conditions, the resulting event representations are equivalent to those derived from sparse event sequences generated with ESIM~\cite{esim}.
While our approach reduces average runtime by 34\% compared to generating event representations with ESIM’s CUDA-based implementation, its primary advantage lies in the significantly improved memory efficiency, illustrated in the appendix.
Compared to ESIM, our method achieves a reduction of more than four times in peak GPU memory usage, removing a key bottleneck for large-scale event simulation.
A more detailed and visual explanation of our event rendering approach are provided in the appendix.

\section{Experiments}
We evaluate our approach in simulation using Flightmare (Sec.~\ref{sec:exp_simulation}) and validate its real-world performance on a quadrotor platform similar to~\cite{Loquercio2021Science} (Sec.~\ref{sec:real_world}), with platform details in the appendix.
\subsection{Simulation}
\label{sec:exp_simulation}
\begin{figure}[t]
    \centering
    \begin{minipage}{0.56\textwidth}
        \centering
        \includegraphics[width=0.99\textwidth]{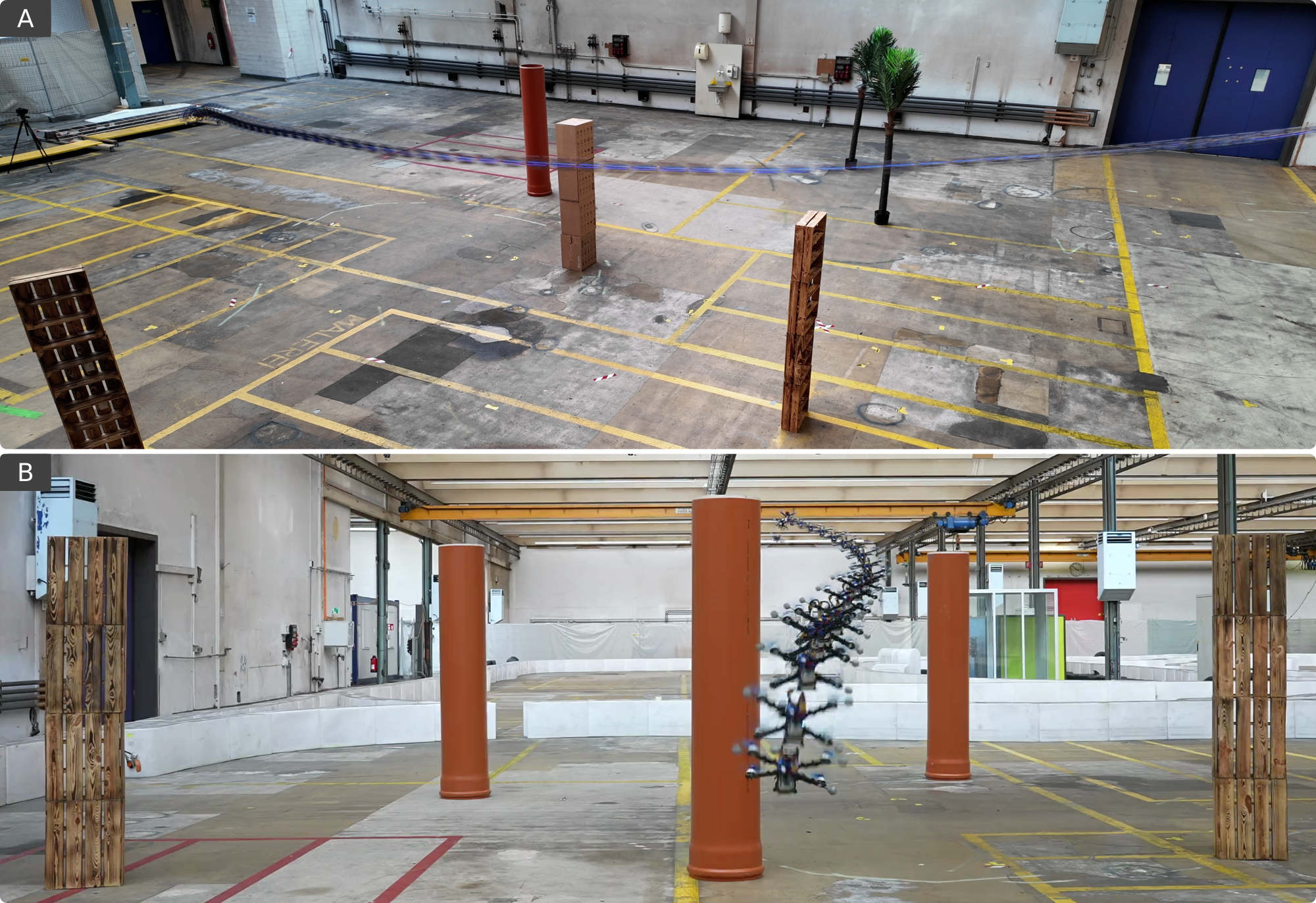}
        \caption{
        Our end-to-end policy demonstrates reliable flight in complex (A/B) real-world environments using only a monocular event camera and onboard compute.
        }
        \label{fig:real_world_images}
    \end{minipage}
    \hfill
    \begin{minipage}{0.42\textwidth}
        \centering
        \includegraphics[width=1.02\textwidth]{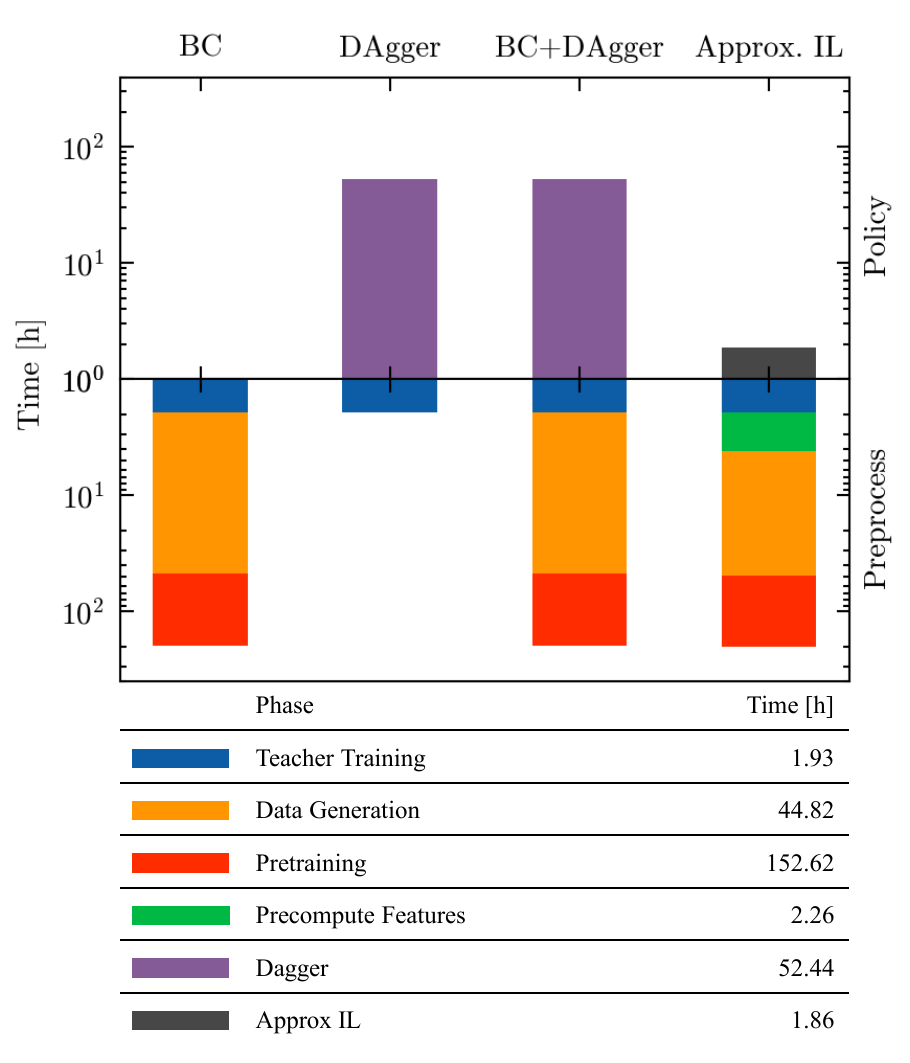}
        \caption{
        Training time grouped into a \textit{Preprocess} and a \textit{Policy} stage.
        }
        \label{fig:training_time}
    \end{minipage}
\end{figure}

\textbf{Setup} 
We construct a simulated forest environment where trees are modeled as tall cylinders with radii ranging from \SI{0.2}{\meter} to \SI{0.5}{\meter}.
These cylinders are placed within a \SI{100}{\meter}$\times$\SI{100}{\meter} world box rendered in a natural outdoor scene, see appendix.
Tree positions are sampled from a Poisson point process, with the Poisson delta controlling the tree density across environments.
For each Poisson delta value (0.04 and 0.05), we sample 10 environments, resulting in 20 test environments.
In each test episode, the quadrotor starts at the center of the left edge of the environment and receives a direction command pointing to the right.
A flight is successful if the quadrotor travels more than \SI{40}{\meter} in the commanded direction without crashing into obstacles or leaving the world box.
\textbf{Results} 
As shown in Tab.~\ref{tab:sim_results}, our proposed behavior fine-tuning via lightweight simulation leads to a 0.2 higher success rate (1.0 vs. 0.8) than Behavior Cloning (BC).
When increasing the tree density to a Poisson delta of 0.05, the performance of our approach decreases from 1.00 to 0.90, while still maintaining a 0.20 advantage over BC.
Despite access to online event rendering, the number of rendering steps used for our approach is insufficient for DAgger~\cite{ross2011icais} to learn a robust mapping from events and state information to control commands.
Moreover, the equal performance of our approximate IL and the BC+DAgger baseline indicates that our approximation with lightweight states closely matches real event renderings, while substantially reducing the training time, as illustrated in Fig.~\ref{fig:training_time}.
Finally, our event student performs comparably to the teacher policy, which relies on highly informative distance maps, confirming that the teacher–student performance gap is small.
\begin{figure}[t]
    \centering
    \begin{minipage}{0.43\textwidth}
        \centering
        \input{floats/figures/fig_representative_plot}
        \caption{
        Illustrative comparison of mean velocity versus success rates reported by baseline methods using absolute depth.
        }
        \label{fig:representative_plot}
    \end{minipage}
    \hfill
    \begin{minipage}{0.55\textwidth}
        \centering
        \input{floats/figures/fig_top_down_real}
        \caption{
        Top-down view of a real-world flight trajectory alongside approximate obstacle locations.
        Fig.~\ref{fig:real_world_images} A shows the first half of this trajectory.
        }
        \label{fig:top_down_real}
    \end{minipage}
\end{figure}

The advantage of our approximate IL method over BC, the best-performing baseline under the same training conditions, is also evident when only onboard inputs are available, i.e., previous actions and commanded direction.
Without access to the quadrotor state, our method improves the success rate by 0.3 compared to BC (0.5 vs. 0.2), as reported in Tab.~\ref{tab:sim_results_wostate}.
Crucially, both approaches rely on the same number of pre-rendered event representations, since our method avoids costly online event rendering by leveraging approximate policy updates in simulation.
\begin{table*}[!t]
\centering
\footnotesize
\caption{
Simulation results for methods using offline (Off.) and/or online (On.) training input.
\label{tab:sim_results}
}
\begin{tabular}{m{2.6cm}C{3.1cm}C{1.2cm}C{1.7cm}C{1.2cm}>{\centering\arraybackslash}m{1.7cm}}
     & & \multicolumn{2}{c}{Poisson Delta: 0.04} & \multicolumn{2}{c}{Poisson Delta: 0.05} \\
    \cmidrule(lr){3-4} \cmidrule(lr){5-6}
    \textbf{Methods} & Training Input & Success Rate & Mean Velocity [$\frac{\mathrm{m}}{\mathrm{s}}$] & Success Rate & Mean Velocity [$\frac{\mathrm{m}}{\mathrm{s}}$] \\
    \midrule
    BC                          & Off. Event                & 0.80 &  8.53 & 0.70 &  8.13 \\
    \grayrow
    DAgger~\cite{ross2011icais} & On. Event                 & 0.30 &  6.66 & 0.10 &  5.72 \\
    BC+DAgger                   & Off. Event + On. Event & 1.00 &  9.00 & 0.90 &  8.47 \\
    \grayrow
    Approx IL (Ours)            & Off. Event + On. State & 1.00 &  8.76 & 0.90 &  8.53 \\
    \hline
    State Teacher (Ours)        & On. State + Dist. Map     & 1.00 &  8.77 & 0.90 &  8.60
\end{tabular}
\end{table*}

\textbf{Ablations}
To validate our design choices, we conduct ablation studies in simulation using an event-based policy with access to state information in environments generated with a Poisson delta of 0.04.
In addition to the success rate, we report the action MSE between the event student and teacher, and between the approximate student and teacher, as well as the L1 distance between the features of the two student encoders $h_S$ and $\hat{h}_S$ (Feature L1).
As shown in Tab.~\ref{tab:align_ablation}, removing the M3ED training stage with the auxiliary distance map prediction task (BC w/o M3ED) significantly reduces the success rate to 0.30 and results in the largest action difference to the teacher.
This observation suggests that the event and approximate students overfit to the offline action distribution.
Adding M3ED during offline training (BC) improves the success rate to 0.80 at the cost of a larger feature difference between the event and approximate students.
Simply using the approximate student during online training to fine-tune the action decoder through online interactions (Approx IL w/o Off. Align.) decreases the success rate to 0.50, despite lower action MSE.
This can be explained by the action decoder leveraging encoder features specific to the approximate student, resulting in separate feature manifolds for the event and approximate students.
Introducing offline alignment during online training mitigates this issue by encouraging the action decoder to remain robust to feature perturbations.
The resulting model (Approx IL) achieves the best overall performance, with a success rate of 1.00 and the lowest action and feature differences.
To assess the alignment between the approximate and event student, we compute the action MSE on events rendered from the approximate student rollout states.
The resulting MSE of 0.00199 is lower than that obtained on teacher rollout states (0.00249), indicating good alignment between the two students.
Additional ablations on learnable parameters, different levels of state information, and the distance map decoder are provided in the appendix.
\begin{table}[t]
    \centering
    \begin{minipage}{0.37\textwidth}
        \centering
\centering
\footnotesize
\begin{tabular}{m{1.5cm}C{0.7cm}>{\centering\arraybackslash}m{1.7cm}}
     & \multicolumn{2}{c}{Poisson Delta: 0.04} \\
    \cmidrule(lr){2-3}
    \textbf{Methods} & Success Rate & Mean Velocity [$\frac{\mathrm{m}}{\mathrm{s}}$] \\
    \midrule
    BC           & 0.20 & 8.08 \\
    \grayrow
    Approx IL    & 0.50 & 8.17
\end{tabular}

        \caption{Simulation results without quadrotor state.}
        \label{tab:sim_results_wostate}
    \end{minipage}
    \hfill
    \begin{minipage}{0.59\textwidth}
        \centering
\footnotesize
\begin{tabular}{m{1.9cm}C{0.8cm}C{1.cm}C{1.cm}>{\centering\arraybackslash}m{1.375cm}}
     & & \multicolumn{2}{c}{Action MSE} & Feature L1 \\
    \cmidrule(lr){3-4} \cmidrule(lr){5-5}
    \textbf{Methods} & Success Rate & Event-Teacher & Approx-Teacher & Event-Approx \\
    \midrule
    BC w/o M3ED                 & 0.30 & 0.0596 & 0.0505 & 0.664 \\
    \grayrow
    BC                          & 0.80 & 0.0543 & 0.0667 & 0.748 \\
    Approx IL w/o Off. Align.   & 0.50 & 0.0248 & 0.0126 & 0.748 \\
    \grayrow
    Approx IL                   & 1.00 & 0.0122 & 0.0108 & 0.568 
\end{tabular}

        \caption{Ablation study for the approximate IL method.}
        \label{tab:align_ablation}
    \end{minipage}
\end{table}

\begin{figure}[ht]
    \centering
    \begin{minipage}[c]{0.55\columnwidth}
        \centering
        \includegraphics[width=\linewidth]{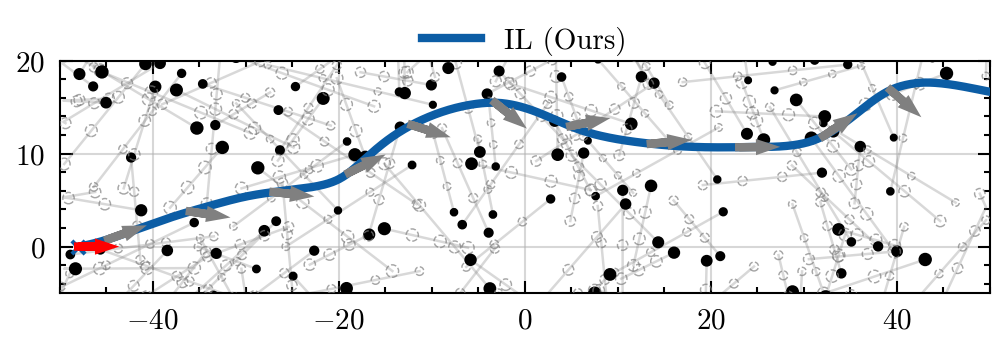}
    \end{minipage}
    \begin{minipage}[c]{0.38\columnwidth}
        \centering
        \footnotesize
        \begin{tabular}{m{1.4cm}C{1.2cm}>{\centering\arraybackslash}m{1.7cm}}
            \textbf{Methods} & Success Rate & Mean Velocity [$\frac{\mathrm{m}}{\mathrm{s}}$] \\
            \midrule
            BC+DAgger   & 0.9 & 8.76  \\
            \grayrow
            Approx IL   & 0.9 & 8.47
        \end{tabular}
    \end{minipage}
\vspace{-8pt}
\caption{Simulation results with dynamic obstacles moving randomly at \SIrange{0.5}{1.5}{\meter\per\second}.}
\label{fig:dynamic_obstacles}
\end{figure}

\textbf{Dynamic Obstacles}
To further assess the generalization of our approximate imitation learning framework, we evaluate our policy and the BC+DAgger baseline in simulation with dynamic obstacles moving in random directions at constant velocities sampled between \SI{0.5}{\meter\per\second} and \SI{1.5}{\meter\per\second}.
Obstacle initial positions are sampled with a Poisson Delta of 0.04.
Without retraining, our policy achieves a success rate of 0.9, as shown in Fig.~\ref{fig:dynamic_obstacles}.
These results demonstrate that the learned policy can reliably avoid moving obstacles despite being trained only on static obstacles.
\textbf{Representative Comparison}
To put the performance of our method into context with existing work, we plot the results reported in related work~\cite{Zhang_2025} in Fig.~\ref{fig:representative_plot} and include the performance of our method for reference.
Although the forest density and evaluation metrics are the same across methods, the underlying simulators and environments differ.
Furthermore, the reported baselines FastPlanner~\cite{zhou2019robust}, Reactive~\cite{florence2020integrated}, Agile~\cite{Loquercio2021Science}, and DiffAgile~\cite{Zhang_2025} use depth information in addition to state information. 
This represents a significant advantage over our monocular event camera setup.
Consequently, this comparison should be interpreted as a representative reference to relate the different success rates and achieved velocities, rather than as a direct one-to-one comparison.
\textbf{Training Time}
Fig.~\ref{fig:training_time} shows the time required by each baseline for the individual training steps, grouped into a \textit{Preprocess} and a \textit{Policy} stage.
The \textit{Preprocess} stage consists of steps that can be reused across multiple policy trainings and therefore do not significantly contribute to the development time for a new policy.
In contrast, the \textit{Policy} stage includes steps that must be executed for every new policy.
Since all evaluated methods rely on IL, they require the training of a teacher policy using lightweight state and perception observations (see appendix), which is inexpensive and takes \SI{1.93}{\hour}.
Except for DAgger, all methods also require \SI{44.82}{\hour} to generate offline event data (Data Generation) and \SI{152.62}{\hour} for the subsequent offline training (Pretraining).
The offline dataset can be reused across any \textit{Policy} stage configurations, while the pretrained network remains reusable as long as the policy architecture is unchanged.
To reduce training time in the \textit{Policy} stage, our proposed approximate IL precomputes event features in \SI{2.26}{\hour} for all offline samples, since its event encoder remains frozen during policy fine-tuning.
The substantial advantage of our approximate IL becomes evident in the \textit{Policy} stage, where the reliance of DAgger methods on online event rendering leads to a training time of \SI{52.44}{\hour}, whereas approximate IL requires only \SI{1.86}{\hour}.
This corresponds to a reduction in policy training time by approximately a factor of 28.
\subsection{Real-World}
\label{sec:real_world}

\textbf{Results}
Our event-based policy is capable of flying reliably through indoor environments with varying levels of clutter using quadrotor state estimates from an external positioning system (VICON), as shown in Fig.~\ref{fig:real_world_images}.
Tested in five environments with obstacle counts that increase from sparse to dense, reaching up to ten obstacles, the quadrotor equipped with a single event camera successfully navigates trajectories through the cylindrical obstacles with a 5/5 success rate.
The policy reaches a peak speed of \SI{9.8}{\meter\per\second} in these indoor tests, as illustrated in Fig.~\ref{fig:top_down_real}, which shows the velocity profile along the trajectory together with the approximate obstacle locations.
This strong performance achieved by an end-to-end event policy is likely supported by the inclusion of real-world event data during the pretraining stage of our approximate IL framework, which helps facilitate transfer from simulation to the real world.
\textbf{Limitations}
Quadrator crashes at high speed with onboard event cameras are costly, preventing real-world evaluation of underperforming baselines.
Even an external safety guard struggled to stabilize the drone at high speeds and provided no protection against obstacle collisions.
Furthermore, while our vision-based end-to-end variant outperforms the best baseline, it still lags behind our approach using external state estimation. 
This gap could be reduced through auxiliary state-prediction tasks during offline training or by leveraging IMU data during offline training while omitting it during online fine-tuning within our approximate IL framework.

\section{Conclusion}
While event cameras offer compelling advantages, such as high temporal resolution and low latency, they remain underused in robotics due to the high computational cost of simulating events.
In this work, we propose an approximate imitation learning framework that significantly reduces the number of required event renderings using offline datasets from simulation and the real world.
To increase the generalization of the policy, we introduce an approximate student that predicts the actions of the event-based student using privileged state information.
Since this state information can be simulated efficiently, the policy can be fine-tuned through online interactions without rendering events.
To further accelerate the validation of event-based policies, we also propose a fully parallelized method for directly generating event representations from high-speed videos.
Experiments in both simulation and the real world demonstrate that our approach enables fast and agile quadrotor flight through cluttered environments using a single event camera, without relying on intermediate representations.
While we apply our approximate IL to event data, the framework naturally extends to other modalities that are expensive to simulate, such as tactile sensors, radar, or LiDAR.

\acknowledgments{
This work was supported by the European Union’s Horizon Europe Research and Innovation Programme under grant agreement No. 101120732 (AUTOASSESS), and the European Research Council (ERC) under grant agreement No. 864042 (AGILEFLIGHT).
}

\clearpage
\section{Appendix}
\subsection{Simulation Environment}
\begin{figure*}[h]
\centering
\includegraphics[width=0.98\textwidth]{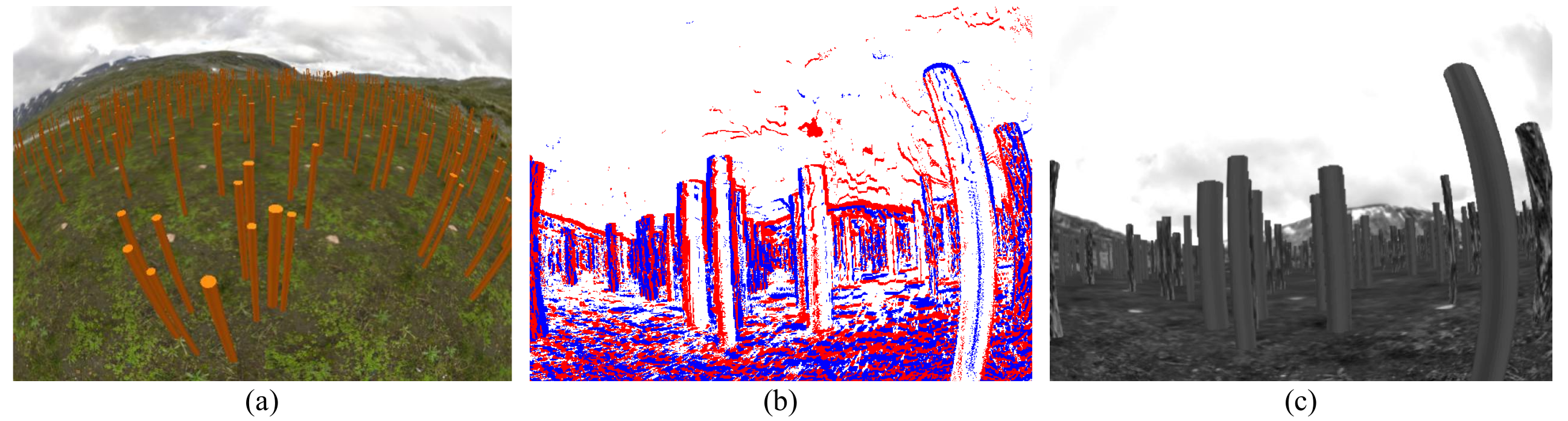}
\caption{
The simulation environment consists of tall cylindrical obstacles placed in a natural scene (a).
The scene is captured by a simulated event camera mounted on the quadrotor (b).
For reference, the corresponding grayscale images are also shown (c).
}
\label{fig:sim_environment}
\end{figure*} 
Figure~\ref{fig:sim_environment} (a) illustrates the simulation environment, which features cylindrical orange obstacles embedded within a natural scene. 
High-framerate grayscale images, shown in Fig.~\ref{fig:sim_environment} (c), are efficiently converted by our vectorized event rendering pipeline into event stacks, as depicted in Fig.~\ref{fig:sim_environment} (b). 
This process preserves fine-grained temporal dynamics while enabling efficient event-based perception.
\subsection{Network Architecture}
\begin{figure*}
\centering
\includegraphics[width=0.9\textwidth]{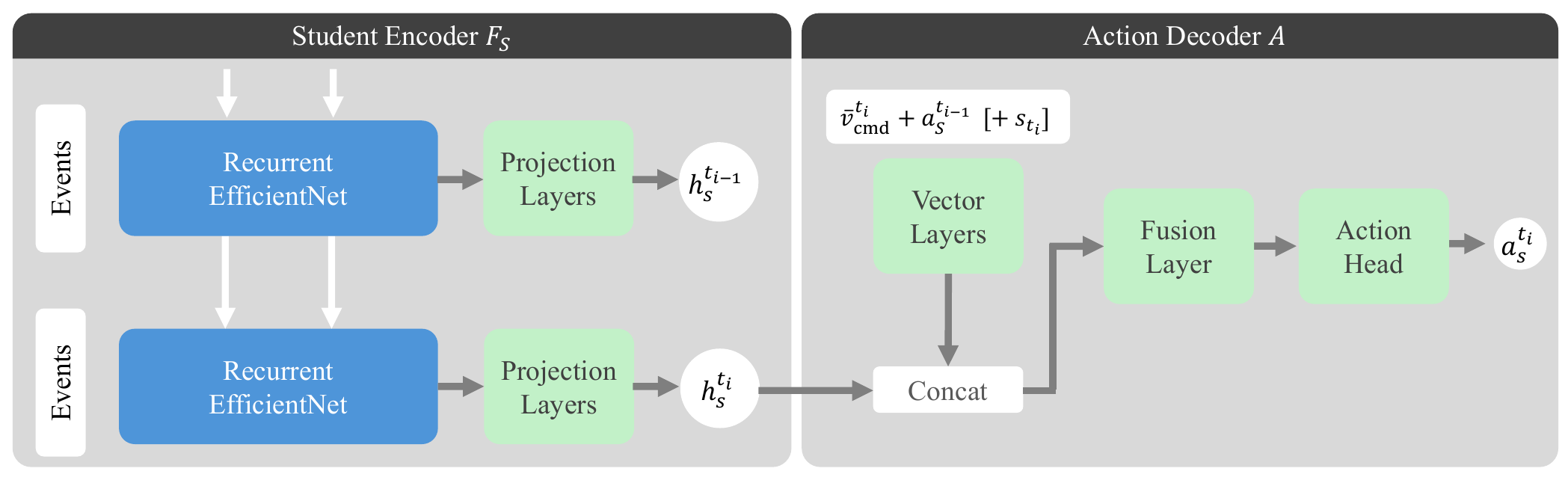}
\caption{
The event student receives events in the form of event representations, which are first encoded into features $h^{t_i}_S$ by the event student encoder $F_S$.
The student encoder $F_S$ consists of a recurrent EfficientNet~\cite{tan2019efficientnet} followed by the \textit{Projection Layers}.
These event features are then concatenated with the outputs of the \textit{Vector Layers}, which encode auxiliary inputs, i.e., the direction command $\bar{v}_{t_i}$, the previous action $a^{t_{i-1}}_S$, and, optionally, the state information $s_{t_i}$.
The combined features are fused in the \textit{Fusion Layer}, and the final actions $a^{t_i}_S$ are produced by the \textit{Action Head}.
}
\label{fig:network}
\end{figure*}

Our framework employs three networks: a teacher, an event-based student, and an approximate student.
The teacher is implemented as a three-layer multilayer perceptron (MLP) with ReLU activations and a final tanh layer, following standard designs used in quadrotor control~\cite{xing2024multi}.
To efficiently process event representations and capture temporal dynamics, the event-based student uses an EfficientNet-B0 encoder~\cite{tan2019efficientnet}, without the final classification layer and with two recurrent GRU layers inserted after the third and fourth stages of the backbone.
The encoded spatial features are flattened through global average pooling and subsequently passed through the \textit{Projection Layers}, which are implemented as a two-layer MLP with ReLU activations.
The encoder of the approximate student is a lightweight three-layer MLP with ReLU activations.
Both students share the same action decoder, which employs a fully connected \textit{Fusion Layer} to merge the event features $h^{t_i}_s$ with the features produced by the \textit{Vector Layers} from auxiliary inputs, namely the direction command $\bar{v}^{t_i}_{\mathrm{cmd}}$, the previous action $a^{t_{i-1}}_S$, and, optionally, the quadrotor state $s_{t_i}$.
The fused features are then further processed by a two-layer \textit{Action Head} with ReLU activations and a final tanh layer.
The processing pipeline of the event student is shown in Fig.~\ref{fig:network}.
Both auxiliary decoders, $D$ and $D_o$, are implemented as four-layer MLPs with ReLU activations.
All networks are trained using the Adam optimizer~\cite{Kingma15iclr}, with a learning rate scheduled via cosine annealing with warm restarts~\cite{sgdr17iclr}.
The input to the event-based student consists of event representations with a resolution of $213{\times}160$, obtained by downsampling the coordinates of an event stream originally at $640{\times}480$.
Importantly, this integer-division downsampling preserves certain high-resolution spatial details that would not be captured by a native $213{\times}160$ event camera, which requires the original event representations to be rendered at $640{\times}480$.
\subsection{Training and Evaluation Details}
\textbf{Training}
Offline training alternates between update steps on M3ED and simulated data.
The offline loss weights are set to $1$ for the distance-map loss, $100$ for the teacher-student action MSE, $1$ for the approximate-student action MSE, and $1$ for the approximate-student feature L1 loss.
During online training, the DAgger action loss has weight $1$, while the offline alignment uses weights of $0.5$ for the student-approximate-student action MSE and $1$ for the corresponding feature L1 loss.
For the offline alignment, we sample 2000 offline samples per DAgger iteration, with the sampled subset varying across iterations.
\textbf{Evaluation}
We use a single training seed and report the mean performance across 10 environments, following the protocol of~\cite{Zhang_2025, Loquercio2021Science}.
For a fair comparison, the online DAgger baselines, DAgger and BC+DAgger, are trained using the same number of rendering steps ($10^6$) as used to generate the simulated offline dataset.
The key difference is that the online DAgger baselines require computationally expensive event rendering for every training run, whereas our approximate IL approach and other offline methods, such as BC, can reuse the offline dataset across runs.
%

\subsection{Event Rendering}
Training end-to-end robot policies with visual inputs in simulation requires a large number of environment steps, which is particularly problematic for event-based policies because training time scales with the cost of simulating events.
This cost is high, as accurate event simulation requires rendering high-frame-rate videos that capture the fine temporal resolution needed to truthfully simulate event sequences.
Each event sequence consists of individual events represented as tuples $(\mathbf{x_k}, p_k, t_k)$, where $\mathbf{x_k}$ is the pixel location, $p_k \in \{+1, -1\}$ is the polarity, and $t_k$ the timestamp.
To simulate events, event generation is commonly approximated using the noise-free model of~\cite{Gallego20pami}.
Specifically, once a high-frame-rate video is synthesized, events are generated by evaluating the change in logarithmic intensity $L=\log(I_{t})$ at a single pixel $\mathbf{x_k}$ over time. 
An event is triggered at timestep $t_k$ if the logarithmic intensity change from the last triggered event at timestep $t_{k-1}$ exceeds a given contrast threshold $C$, as shown in Eq.~\ref{eq:event_generation}.
\begin{align}
    \Delta L(\mathbf{x_k}, t_k) = L(\mathbf{x_k}, t_{k-1}) - L(\mathbf{x_k}, t_{k}) = p_k C.
\label{eq:event_generation}
\end{align}
Here, the polarity $p_k$ indicates the direction of the intensity change.
In popular video-to-event methods~\cite{Gehrig20cvpr, esim}, the frame rate of the underlying video is chosen such that the maximum optical flow between two frames does not exceed one pixel.

\begin{figure}[ht]
\centering
\includegraphics[width=0.7\textwidth]{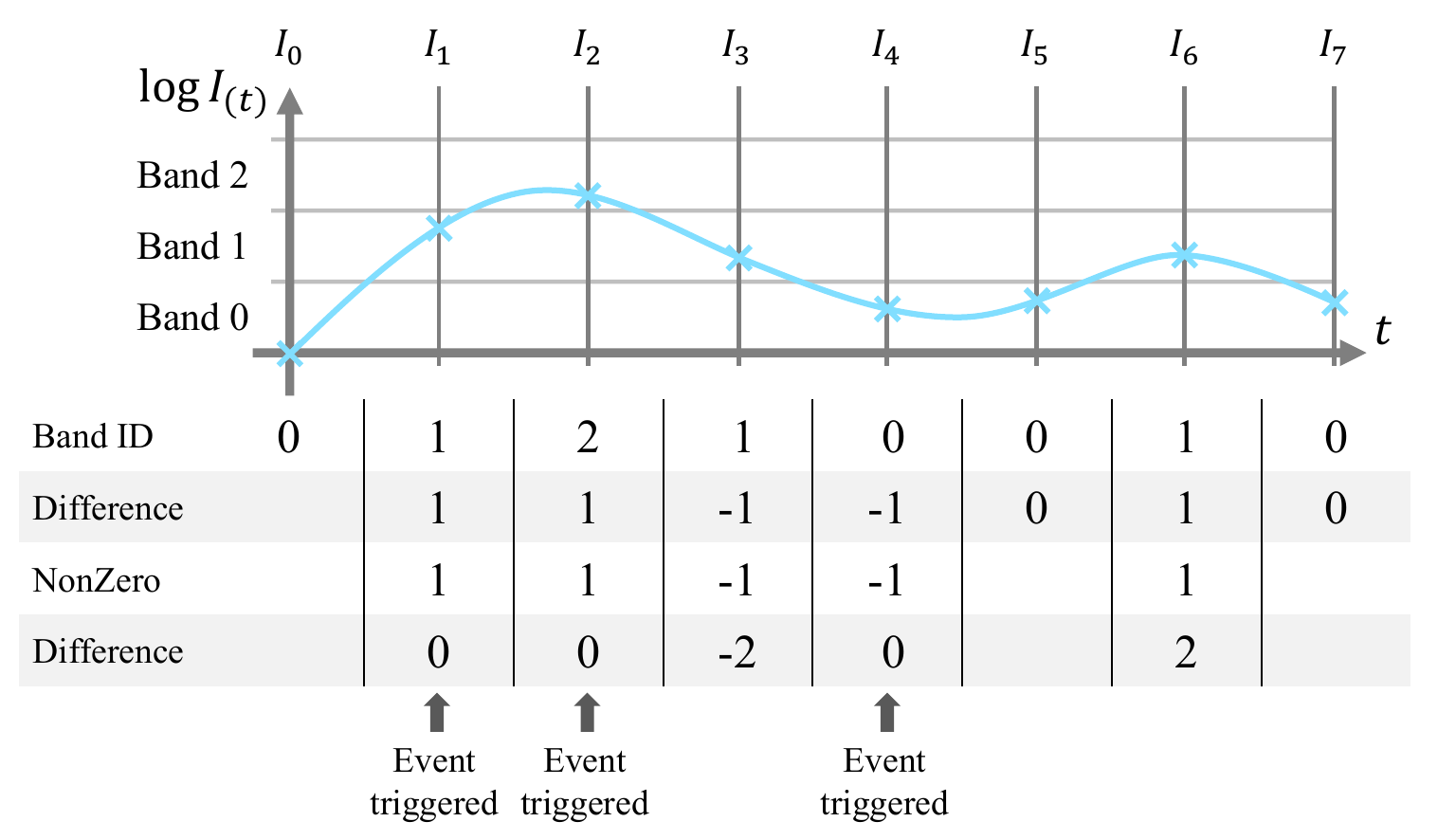}
\caption{
Our vectorized event generation first quantizes log intensities into bands and computes band differences between adjacent timesteps.
Neighboring non-zero values are then differenced again. 
Events are triggered where this second subtraction yields zero.
}
\label{fig:event_generation}
\end{figure} 
In our use case, the event stream is ultimately converted into a dense spatiotemporal representation, which is then processed by a lightweight neural network.
Thus, generating a full event sequence only to later transform it into a dense representation introduces significant and unnecessary computational overhead.
Inspired by~\cite{bhattacharya2024monocular}, we follow the idea of directly generating event representations, effectively bypassing the intermediate step of synthesizing sparse event sequences.
This design has two main advantages: (i) the direct generation of event representations reduces computation time and memory usage by avoiding explicit event sequence simulation, and (ii) it removes the need for high temporal fidelity, as the event representation uses a coarse discretization of the time dimension.
In contrast to~\cite{bhattacharya2024monocular}, which generates single-channel representations, our proposed approach truthfully constructs a multi-channel event tensor that accumulates the events in a given temporal window for each polarity~\cite{mostafavi2021learning}.
Importantly, our method generates event representations for multiple videos in parallel using vectorized operations, significantly accelerating the simulation.

\begin{figure}[!t]
\centering
\includegraphics[width=0.55\textwidth]{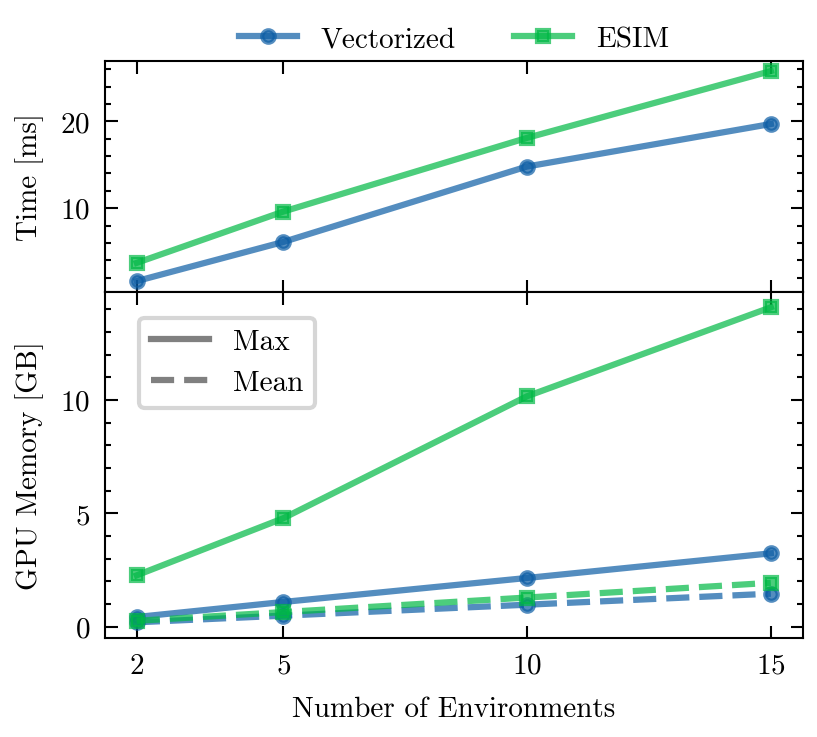}
\caption{
The maximum and mean GPU memory usage and the required computation time are shown for our vectorized event generation (Vectorized) and ESIM’s GPU-based event generation (ESIM) across different numbers of environments.
Our method leads to a 34\% reduction in mean runtime (top) and significantly lowers peak GPU memory usage (bottom).
}
\label{fig:event_memory}
\end{figure}

The core idea behind our event representation method is to leverage fast, vectorized operations, as illustrated in Fig.~\ref{fig:event_generation}, which shows the log-intensity curve over time alongside the key processing steps.
Given a high-frame-rate video with images $I_0, I_1, ... I_7$, we first compute the log-intensity for each frame and subtract the intensity of the initial frame $I_0$, establishing it as the reference.
In the next step, each normalized log-intensity frame is floor-divided by the contrast threshold $C$, assigning each pixel at each timestep to a discrete \textit{contrast band} based on its current intensity level.
To detect band transitions, we compute the temporal difference of band indices between consecutive frames. 
Steps where the band index remains unchanged, i.e., the difference equals zero, indicate no threshold crossing and are discarded.
Crucially, an event is only triggered when the intensity curve exits the current band in the opposite direction from which it previously entered.
To identify this, we compute a second-order difference of the band indices over time, using only frames with valid, non-zero first-order differences.
An event is triggered at timesteps where this second difference is zero, indicating that the logarithmic difference exceeded the contrast threshold $C$.
The polarity of the triggered event is determined from the sign of the first band ID difference.
To obtain the final event representation, a simple reshaping and summing along a given axis is required.
To account for multiple events being triggered at a given pixel within a single timestep, we take the absolute value of the band ID difference in the first differencing step, which corresponds to the number of threshold crossings.
Additionally, to simulate a non-zero initial reference value, a constant offset can be subtracted from the log-intensity at the normalization stage, allowing for flexible initialization of the reference state.
Our proposed event generation method enables the parallel generation of event representations for multiple video streams through efficient vectorized operations implemented in PyTorch~\cite{paszke2017automatic}.
Except for a few corner cases related to specific initial conditions, the resulting event representations are numerically equivalent to those obtained by first generating full event sequences with ESIM~\cite{esim} and subsequently converting them to dense representations.
While our approach reduces average runtime by 34\% compared to generating event representations with ESIM’s CUDA-based implementation, its primary advantage lies in the significantly improved memory efficiency, illustrated in Fig.~\ref{fig:event_memory}.
Compared to ESIM, our method substantially reduces the maximum GPU memory usage, which ultimately determines whether event simulations at scale are feasible.
Consequently, ESIM fails to scale to higher resolutions and larger numbers of environments, restricting the comparison in Fig.~\ref{fig:event_memory} to half the final resolution ($240\times320$) and at most 15 environments.
In contrast, our method supports large-scale parallel simulation environments, making it well-suited for training event-based control policies at scale.

\subsection{Teacher Training}
The teacher policy is trained using Proximal Policy Optimization (PPO)\cite{schulman2017proximal} within the Flightmare simulator~\cite{song2020flightmare}.
Its observation space includes the commanded direction in the local frame, the z-position of the quadrotor, orientation represented as a flattened rotation matrix, linear and angular velocities, and the previous action command.
To perceive obstacles, the teacher also receives an angular distance map that encodes the closest obstacle distances across 10 angular bins.
Each bin spans \SI{11.25}{\degree}, covering a 120-degree horizontal field of view.
The episode terminates if the quadrotor leaves the simulation boundary, collides with an obstacle, or reaches the maximum number of steps.
The teacher is trained for 8000 PPO update steps, with each update performed on rollouts of 250 steps collected from 100 parallel environments.

The total reward $r_t$ combines several weighted components, with their respective factors $\lambda_{*}$ omitted in Eq.~\ref{eq:reward_quad} for clarity:
\begin{equation}
    r_{\mathrm{t}} = r_{\mathrm{prog}} + r_{\mathrm{act}} + r_{\mathrm{br}} + r_{\mathrm{perc}} + r_{\mathrm{obs\_dist}} + r_{\mathrm{crash}}
    \label{eq:reward_quad}
\end{equation}
{\small
\begin{equation}
    \begin{aligned}
        r_{\mathrm{prog}}  &= \tanh\left({ \frac{\bar{v}_{\mathrm{quad}} \cdot \bar{v}_{\mathrm{cmd}}}{\lVert \bar{v}_{\mathrm{cmd}}\rVert} + 1}\right) 
        \\
        &\cdot \tanh\left({\lVert \bar{v}_{\mathrm{cmd}} \rVert- \lVert \bar{v}_{\mathrm{quad}} - \bar{v}_{\mathrm{cmd}} \rVert} + 1 \right)
        \\ & \cdot \frac{1}{4} \min \left( \frac{\lVert\bar{v}_{\mathrm{quad}}\rVert}{\lVert\bar{v}_{\mathrm{cmd}}\rVert}
        , 1\right)  - \left|v^{z}_{\mathrm{quad}}\right|, \\
        r_{\mathrm{act}} & = -\lVert \bar{a}_{t_i} - \bar{a}_{t_{i-1}}\rVert, \\
        r_{\mathrm{br}} & = -\lVert\bar{a}_{\omega}\rVert, \\
            r_{\mathrm{perc}} & =  \frac{\bar{v}_{\mathrm{cmd}} \cdot \bar{d}_{\mathrm{heading}}}{\lVert \bar{v}_{\mathrm{cmd}}\rVert \lVert \bar{d}_{\mathrm{heading}}\rVert}, \\
        r_{\mathrm{obs\_dist}} &= - \frac{1}{N_{\mathrm{obs}}} \sum_i^{N_{\mathrm{obs}}} \exp\left({-d_{\mathrm{obs}_i}-(\phi_{\mathrm{heading}}-\beta_{\mathrm{obs}_i})^2}\right) \\
        r_{\mathrm{crash}} =& -\lVert\bar{v}_{\mathrm{quad}}\rVert - 1 \quad\text{if crash ceiling, ground or obstacle},
    \end{aligned}
    \label{eq:reward}
\end{equation}
}
where $r_{\mathrm{prog}}$ encourages the quadrotor velocity $\bar{v}_{\mathrm{quad}}$ to align with the commanded direction $\bar{v}_{\mathrm{cmd}}$ to promote goal-directed flight, 
$r_{\mathrm{act}}$ penalizes large changes in consecutive actions $\bar{a}_{t_{i}}$ to ensure smooth control, 
$r_{\mathrm{br}}$ discourages high body-rate actions $\bar{a}_{\omega}$ to support stable flight dynamics, 
$r_{\mathrm{perc}}$ rewards alignment of the yaw direction of the quadrotor $\bar{d}_{\mathrm{heading}}$ with the commanded direction,
$r_{\mathrm{obs\_dist}}$ exponentially penalizes the distance $d_{\mathrm{obs}_i}$ to nearby obstacles $N_{\mathrm{obs}}$ inside the field of view $(\phi_{\mathrm{heading}}-\beta_{\mathrm{obs}_i})$,
and $r_{\mathrm{crash}}$ applies a penalty upon collisions or when the quadrotor exits the simulation boundaries.
Overall, the proposed teacher formulation yields a stable state-based policy capable of reliably flying in highly cluttered environments.
\subsection{Simulated Flight Trajectories}
\begin{figure*}[!t]
\centering
\includegraphics[width=0.95\textwidth]{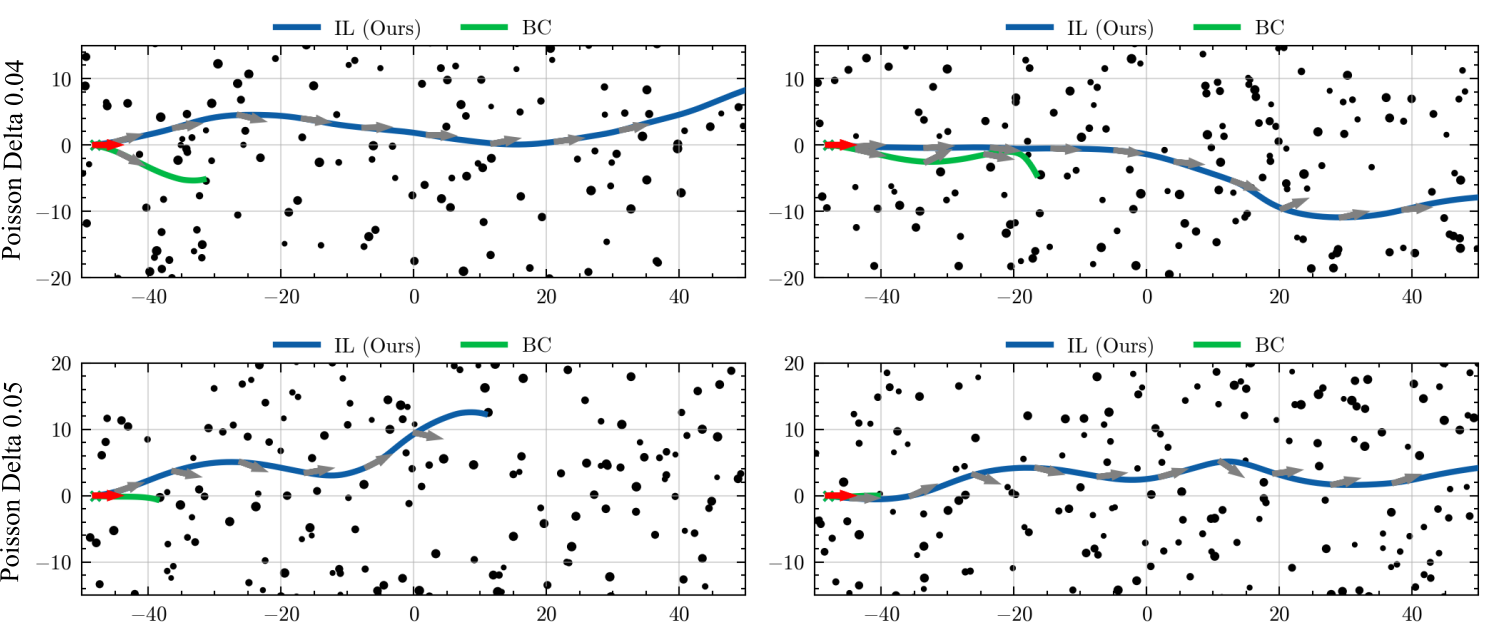}
\caption{
Our proposed method yields a policy that consistently flies longer trajectories in simulation than BC, the closest baseline applicable in the same training setting.
}
\label{fig:sim_trajs_wstate}
\end{figure*} 
Example flight trajectories in four evaluation environments are visualized in Fig.~\ref{fig:sim_trajs_wstate}, comparing our proposed approximate imitation learning method against behavioral cloning (BC), the best-performing baseline trained under identical conditions.

\subsection{Ablations}
To validate our design choices, we conduct several ablation studies in simulation using the standard model, i.e., the event-based policy with access to state information.
The evaluation environments are generated with a Poisson delta of 0.04.
\subsubsection{Learnable Network Parameters}
In this ablation, we examine which components of the network should remain trainable during approximate IL, as summarized in Tab.~\ref{tab:train_ablation}.
In the default configuration, we fine-tune the \textit{Vector Encoder} and the \textit{Fusion Layer}, which leads to the longest mean flight distance before a crash.
When all parameters of the action decoder are updated (Full Action Decoder), the policy still reaches \SI{94.1}{\meter} and maintains a high speed of \SI{9.14}{\meter\per\second}.
Reducing further the number of trainable parameters, either by updating only the biases (Only Biases) or by updating only the final layer (Last Layer), results in slower flight speeds and significantly shorter flight distances, e.g., \SI{46.67}{\meter} when updating only the last layer.
This degradation can be explained by the reduced adaptability of the policy that results from limiting the set of trainable parameters.
Overall, the best performance is achieved when updating the \textit{Vector Encoder} and the \textit{Fusion Layer}, which suggests that there is an optimal trade-off between adapting the policy during the approximate IL phase and preserving the knowledge learned during the pretraining stage.
\subsubsection{Quadrotor State Observations}
To assess the effect of the quadrotor state observations, we ablate policies that receive varying levels of state information, as reported in Tab.~\ref{tab:state_ablation}.
Naturally, providing full state information leads to the highest robustness with a success rate of 1.0.
The removal of angular velocities (w/o Ang. Vel.) has little impact, as the resulting policy still achieves a perfect success rate.
In contrast, excluding translational velocities (w/o Transl. Vel.) reduces the success rate to 0.8 while increasing the mean flight speed to \SI{12.91}{\meter\per\second}, reflecting the difficulty in controlling the speed without absolute velocity observations.
The differing impact of removing angular versus translational velocities can be explained by the ability of the event camera to infer rotational motion through its high temporal resolution, whereas the absolute scale information required to estimate translational velocity is inherently missing in monocular sensing.
When both angular and translational velocities are removed (w/o Vel.), the success rate drops further to 0.7, approaching the performance obtained without any state information (w/o State), which reaches 0.5.
\begin{table}[t]
    \centering
    \begin{minipage}{0.55\textwidth}
        \centering
        \centering
\footnotesize
\begin{tabular}{m{3.1cm}C{1.7cm}>{\centering\arraybackslash}m{1.7cm}}
    \textbf{Methods} & Mean Distance [$\mathrm{m}$] & Mean Velocity [$\frac{\mathrm{m}}{\mathrm{s}}$]\\
    \midrule
    Only Biases             & 76.60 & 7.76 \\
    \grayrow
    Last Layer              & 46.67 & 8.44 \\
    Full Action Decoder     & 94.10 & 9.14 \\
    \grayrow
    Vector+Fusion (Default) & 110.3 & 8.76
\end{tabular}

        \caption{Ablation of trainable action decoder parameters.}
        \label{tab:train_ablation}
    \end{minipage}
    \hfill
    \begin{minipage}{0.43\textwidth}
        \centering
\centering
\footnotesize
\begin{tabular}{m{2.1cm}C{1.0cm}>{\centering\arraybackslash}m{1.7cm}}
    \textbf{Methods} & Success Rate & Mean Velocity [$\frac{\mathrm{m}}{\mathrm{s}}$]\\
    \midrule
    w/o State           & 0.50 & 8.17 \\
    \grayrow
    w/o Vel.            & 0.70 & 11.55 \\
    w/o Transl. Vel.    & 0.80 & 12.91 \\
    \grayrow
    w/o Ang. Vel.    & 1.00 &  9.04 \\
    w State             & 1.00 &  8.76
\end{tabular}

        \caption{Ablation of varying state information levels.}
        \label{tab:state_ablation}
    \end{minipage}
\end{table}
\subsubsection{Distance Map Estimation}
\begin{figure*}[!t]
\centering
\includegraphics[width=0.95\textwidth]{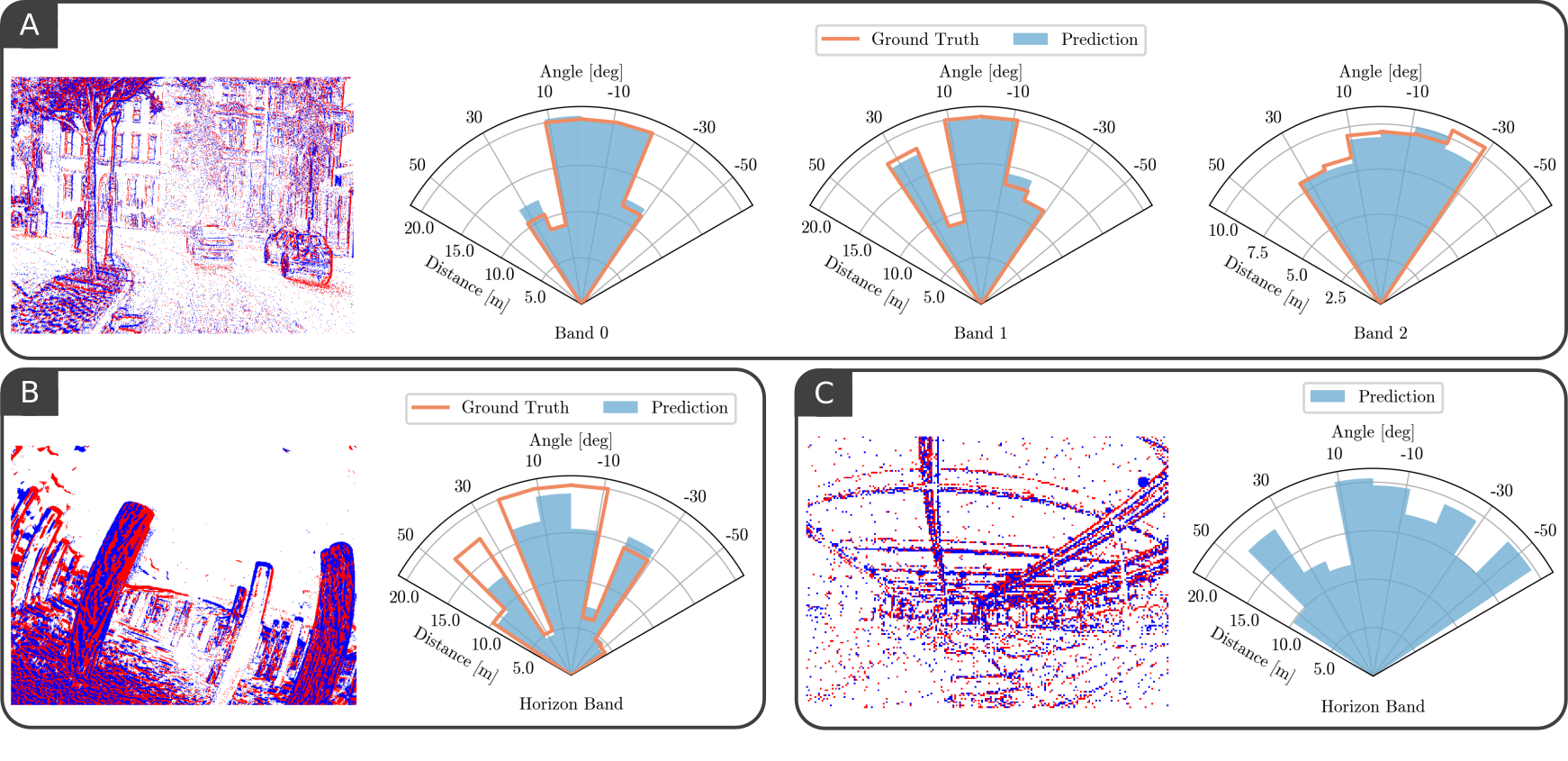}
\caption{
The ground truth distance maps and corresponding predictions are shown for the M3ED dataset (A), simulation data (B), and the real-world test environment (C).
Importantly, the distance maps and associated auxiliary decoders are used only during pretraining and are not involved in action inference of the event student.
}
\label{fig:dist_decoder}
\end{figure*}

To provide insight into the angular distance map estimation, we visualize sample predictions alongside the corresponding ground truth in Fig.~\ref{fig:dist_decoder}.
Importantly, the distance maps and associated auxiliary decoders $D$ and $D_o$ are used only during the pretraining phase and are discarded during action inference of the event student.
In Fig.~\ref{fig:dist_decoder} (A), we show the three ground truth distance bands together with the corresponding predictions for the M3ED dataset.
Even though this sample comes from a hold-out validation split, the auxiliary decoder $D$ accurately predicts the angularly binned distances in an automotive scene.
On simulation data, the auxiliary decoder $D_o$ reliably predicts gravity-aligned angular distance maps, as illustrated in Fig.~\ref{fig:dist_decoder} (B).
Finally, Fig.~\ref{fig:dist_decoder} (C) shows a prediction of a gravity-aligned angular distance map in the real-world test environment onboard a hovering quadrotor.
These qualitative examples demonstrate that the auxiliary decoders can robustly predict angular distance maps using only events, providing evidence that the feature space of the event encoder captures meaningful distance information.

\subsection{Hardware Setup}
\begin{figure}[!t]
\centering
\includegraphics[width=0.55\textwidth]{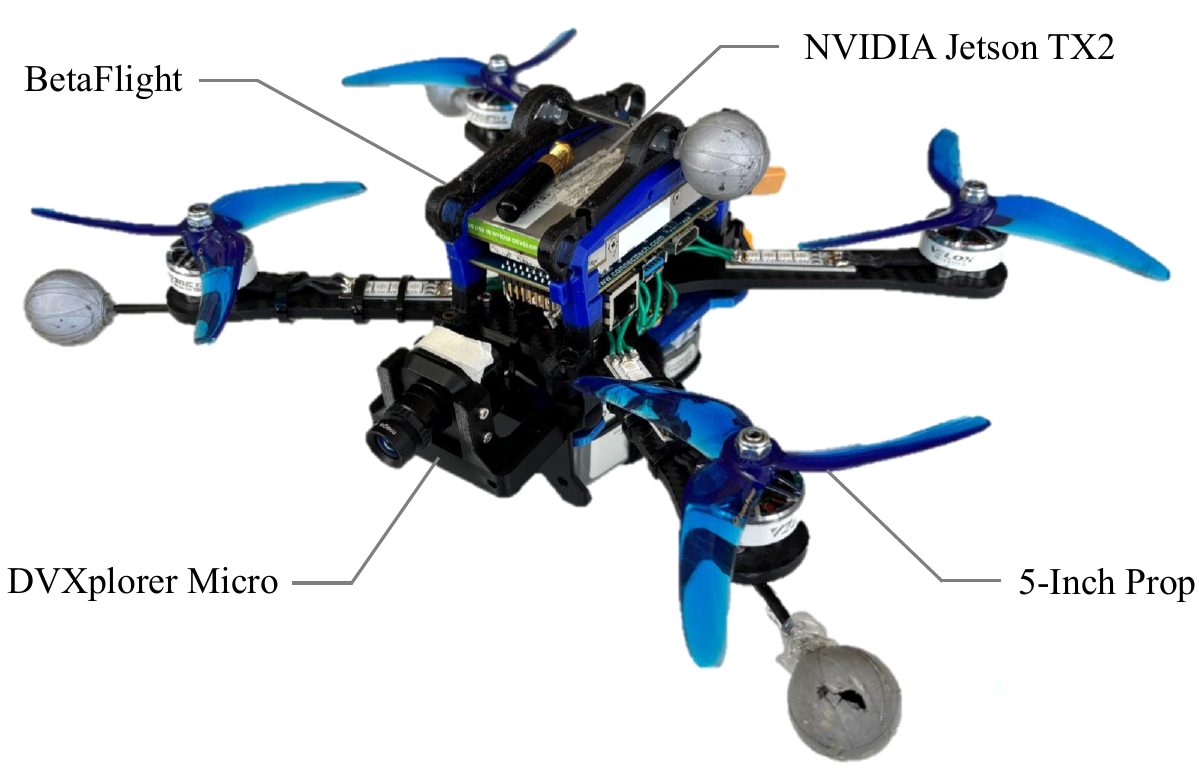}
\caption{
The platform used in the real-world experiments features an NVIDIA Jetson TX2 running the event-based student, which processes events from a DVXplorer Micro sensor.
}
\label{fig:hardware_setup}
\end{figure}

We validate our approach in the real world using a quadrotor platform and onboard computational resources similar to those used in~\cite{Loquercio2021Science}, as shown in Fig~\ref{fig:hardware_setup}.
The quadrotor is equipped with Hobbywing XRotor 2306 motors and 5-inch propellers, resulting in a total weight of approximately \SI{780}{\gram}.
To obtain an accurate drone model in simulation, the inertial parameters and thrust mapping are identified, while selected Flightmare parameters are randomized to account for sim-to-real differences.
Onboard computation is handled by an NVIDIA Jetson TX2, mounted on a ConnectTech Quasar carrier board.
Our neural network outputs control commands in the form of collective thrust and angular rates, which are converted to individual rotor commands using the BetaFlight flight controller.
For event-based perception, we use the DVXplorer Micro sensor with a VGA resolution of $640{\times}480$ pixels, weighing approximately \SI{16}{\gram} without a lens.
Although the DVXplorer Micro includes an onboard IMU, we do not use its inertial measurements in our approach.
Using TensorRT, a forward pass of our event network processing $213{\times}160$ event representations takes on average \SI{11.6}{\milli\second} on the Jetson TX2, providing sufficient margin to meet the \SI{20}{\milli\second} requirement of the \SI{50}{\hertz} control frequency.
\subsection{Extension Beyond Event Cameras}
Our approximate imitation learning framework assumes that offline training establishes a shared task-relevant latent space between the approximate student and student, preserving factors such as motion, obstacle distance, and free space while discarding event-camera-specific appearance or lighting cues.
Extending the framework to other sensor modalities, therefore, requires the approximate student and student observations to share sufficient task-relevant information, while allowing sensor-specific information to be removed during alignment.
If such shared information is absent, the offline alignment can not establish a common task-relevant representation, and the online training phase with the approximate student may degrade student performance.
In this case, however, the observations themselves may provide insufficient information to solve the task.
%

\bibliography{references}  

\end{document}